\pdfoutput=1
\documentclass[11pt]{article}
\usepackage{ACL2023}
\usepackage{times}
\usepackage{latexsym}
\usepackage[T1]{fontenc}
\usepackage[utf8]{inputenc}
\usepackage{microtype}
\usepackage{inconsolata}
\usepackage{graphicx}

\title{Tagengo: A Multilingual Chat Dataset}

\author{Peter Devine \\
  Lightblue KK. (Tokyo, Japan) \\
  \texttt{peter@lightblue-tech.com}
}
\begin{document}
\maketitle
\begin{abstract}

Open source large language models (LLMs) have shown great improvements in recent times. However, many of these models are focused solely on popular spoken languages. We present a high quality dataset of more than 70k prompt-response pairs in 74 languages which consist of human generated prompts and synthetic responses. We use this dataset to train a state-of-the-art open source English LLM to chat multilingually.
We evaluate our model on MT-Bench chat benchmarks in 6 languages, finding that our multilingual model outperforms previous state-of-the-art open source LLMs across each language. We further find that training on more multilingual data is beneficial to the performance in a chosen target language (Japanese) compared to simply training on only data in that language. These results indicate the necessity of training on large amounts of high quality multilingual data to make a more accessible LLM.

\end{abstract}

\section{Introduction}

Recently, open source large language models (LLMs) have grown drastically in both popularity and performance. Models such as Llama 3~\cite{llama3modelcard} have exceeded the performance of previous state-of-the-art proprietary models like GPT3.5~\cite{ouyang2022training} on popular robust benchmarks including the Chatbot Arena leaderboard~\cite{chiang2024chatbot}. These open source LLMs are also increasingly being used in commerical AI chat products such as the Meta AI assistant~\cite{metaai}.

However, many current LLMs exhibit lower performance on languages outside of English~\cite{achiam2023gpt}. Indeed, Llama 3 itself is currently an English-only LLM, meaning that even when it is prompted in a language besides English, it often replies in English. This limits the potential user base of these LLMs due to the fact that less than 1.5 billion of the world's more than 8 billion population can speak English~\cite{cia2021worldfactbook,ethnologue2024}.
Therefore, we set out to train a state-of-the-art open source LLM (Llama 3) to be able to chat not only in English, but in many languages.

In order to make English-focused LLMs accessible in other languages, previous work has fine-tuned these models on non-English data~\cite{elyzallama2023,sengupta2023jais,damonlpsg2023seallm}.

Many multilingual chat datasets such as MultiAlpaca~\cite{wei2023polylm} and Aya~\cite{singh2024aya} cover many languages and tasks but can also lack natural prompts and high quality responses.

For this reason, we created a large, diverse, high quality multilingual dataset using more than 70k human generated prompts in 74 languages and generated responses from these using state-of-the-art proprietary chat models.
We used this dataset to train two models, a multilingual LLM and a Japanese-only LLM, both supervised fine-tuned models based on the Llama 3 8B Instruct model.

We found that our model achieved better evaluation scores on multilingual chat benchmarks compared to the similarly sized state-of-the-art open source models, indicating the high quality and diversity of our training dataset. We also find that our multilingual-trained LLM performs better on Japanese chat benchmarks compared to our Japanese-only-trained LLM, indicating that transfer learning from training on other languages is beneficial for training even monolingual models outside of English.

Our findings combine to inform the community of exactly how to fine-tune monolingual LLMs to create a strong multilingual model. 

We make our training data\footnote{\url{https://huggingface.co/datasets/lightblue/tagengo-gpt4}}, training code\footnote{\url{https://github.com/lightblue-tech/suzume}}, and trained models\footnote{\url{https://huggingface.co/lightblue/suzume-llama-3-8B-multilingual}}\footnote{\url{https://huggingface.co/lightblue/suzume-llama-3-8B-japanese}} publicly available for free use online.

\begin{figure}[h]
    \centering
    \includegraphics[width=200pt]{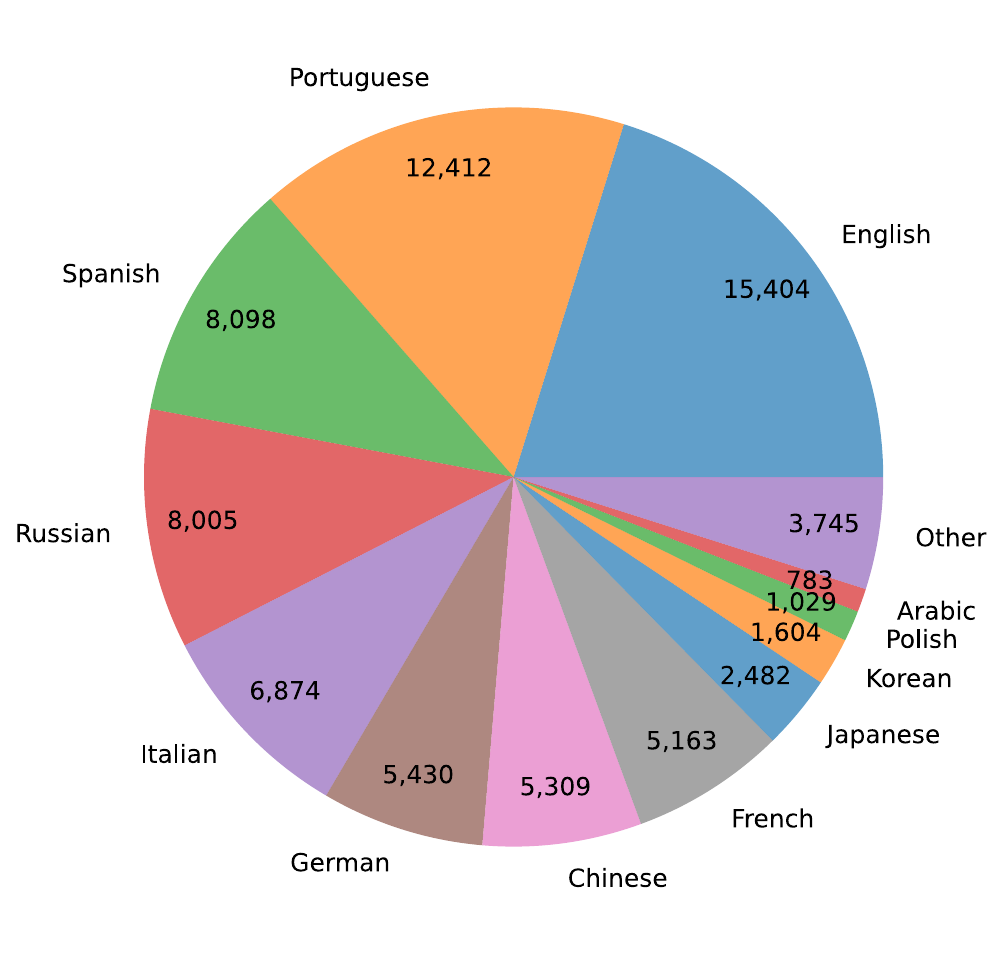}
    \caption{Distribution of the languages found in the Tagengo dataset}
    \label{fig:language distribution}
\end{figure}


\section{Related Work}

In the literature, strong foundation models such as Llama 2~\cite{touvron2023llama} and Gemma~\cite{team2024gemma} have been subsequently fine-tuned on data from a specific language or languages, including Japanese~\cite{elyzallama2023}, Arabic~\cite{sengupta2023jais}, and South-East Asian languages~\cite{damonlpsg2023seallm}.
Fine-tuning has often shown to improve the accuracy of the resultant LLM on tasks in that language. However, the training dataset of these models are often not shared, making it difficult to create a truly multilingual LLM across many languages.

Some multilingual chat datasets do exist that can be used for training LLMs. 
MultiAlpaca~\cite{wei2023polylm} is a multilingual dataset of 133K prompt-response pairs covering 11 languages that were generated in a similar manner to Alpaca~\cite{taori2023alpaca}. This dataset was created by generating synthetic prompts from a small number of English seed prompts and then answering these prompts using an large-scale LLM, GPT3.5~\cite{ouyang2022training}. Because these prompts are generated synthetically, this data may not reflect the sorts of prompts that real users may use with an LLM, potentially limiting the ability of models trained on this data to be used practically. Moreover, the prompts and responses for this dataset were generated using GPT3.5, meaning that the quality of the data may not be as high as if a state-of-the-art LLM was used, like GPT4~\cite{achiam2023gpt}.

xP3 (Crosslingual Public Pool of Prompts) ~\cite{muennighoff2022crosslingual} is a dataset of more than 78 million examples covering 46 languages. This dataset was generated by templating other datasets (e.g. translation datasets, classification datasets) into a prompt-response forma. While this dataset is large, the templating process limits the usefulness of the dataset as it results in prompts that are not necessarily similar to what an actual user of an LLM would ask. The templating process can also result in unnatural answers, with single word answers being given where a fuller answer may be more appropriate from an LLM.

Aya~\cite{singh2024aya} is a dataset of 204k human-annotated prompt-completion pairs covering 65 languages. The majority of this dataset was generated by first translating and templating datasets from various languages, which were then corrected and annotated by human labellers. While the human labelling process will prevent as many unnatural utterances enter the dataset, the templating of datasets means that the prompts will still not necessarily be the kind of prompts that an end-user of LLMs would use. Hence, the usefulness of this dataset in training multilingual LLMs is limited by its data-generation process.

The ShareGPT dataset used by models such as Vicuna~\cite{chiang2023vicuna} and OpenChat~\cite{wang2023openchat} contain approximately 70k open source conversations between users and GPT3.5~\cite{ouyang2022training} and 6k conversations between users and GPT4~\cite{achiam2023gpt}, meaning that the prompts used in these datasets are often much more naturalistic to a real LLM use-case. However, the majority of these prompts are in English, meaning that this dataset is limited in its use in training multilingual models. Moreover, due to the fact that that majority of this dataset contains data generated from GPT3.5, its usefulness in training is limited as many other models have now surpassed the performance of GPT3.5 in English~\cite{starling2023,llama3modelcard}. The amount of multilingual data in the higher quality GPT4 subset of the ShareGPT dataset is small, meaning that its usefulness in training is constrained by its size.

To address the shortcomings in existing public datasets, we created a large, diverse, high quality multilingual dataset using more than 70k human generated prompts in 74 languages and generated responses from these using state-of-the-art proprietary chat models.

\section{Method}

In this section, we detail how we generated our training dataset, our training method, and finally our evaluation techniques.

\subsection{Tagengo Dataset Creation}

First, to generate our dataset, we sampled prompts from the million row LMSYS-Chat-1M dataset~\cite{zheng2023lmsyschat1m}. These prompts were collected from users speaking to one of 25 LLMs on the Vicuna demo and Chatbot Arena website\footnote{\url{https://chat.lmsys.org/}}.

We cleaned this dataset by first removing all prompts which contains an OpenAI Moderation Endpoint\footnote{\url{https://platform.openai.com/docs/guides/moderation}} flag in order to remove explicit, sexual, or illegal material. 

We then removed all prompts which were listed as a non-recognised or fictional language (unknown, Klingon, xx, zp, and zzp). 

We removed any prompts which contained the string ``name'' when lower-cased, as NAME0, NAME1 etc. was used as the placeholders for anonymised material. Effectively, this removed any anonymised prompts from our dataset.

We then removed any prompts which contained the following keywords: ``gpt'', ``vicuna'', ``alpaca'', ``llama'', ``koala'', ``claude'', ``guanaco''. This was done to remove prompts which referred explicitly to the LLMs that were being tested in the Chatbot Arena as many prompts asked about the LLM specifically, which we theorize is less useful in a more general context. 

We then used the FastText~\cite{joulin2016fasttext} LangDetect library\footnote{\url{https://github.com/zafercavdar/fasttext-langdetect}} to determine the confidence level of classifying a particular language. We filtered out all prompts in which the confidence level of the language indicated in the original LMSYS-Chat-1M paper was less than 80\%. This was done to filter out ambiguous language examples, as we later sample per-language.

Finally, we analysed the number of tokens of both the first prompt and LLM response, and removed any prompts in which the combined token total of the first prompt and LLM response amounted to more than 512 tokens. This was done to prevent very long prompts or prompts which elicited very long responses being used in our dataset in order to minimise costs when generating data with these prompts using GPT4.

We then sampled a maximum of 25,000 prompts from each language, which effectively meant we sampled the English prompts in this dataset as only English (380,138) had more than 25,000 examples, while the next most popular language Chinese (21,057) had less than 25,000. This was done to counteract the outweighed prevalence of English within this dataset.

For each language, we then embedded each prompt using the BGE M3 embedding model~\cite{chen2024bge}, which is a state-of-the-art embedding model that supports more than 100 languages. We then compared the prompt embeddings pairwise using the dot product to obtain a similarity score for each prompt pair. We perform fuzzy deduplication by removing one of any prompt pairs which have a similarity score of greater than 0.8 in order to bolster the diversity of our dataset. The amount of data removed from each language varied widely with languages such as Chinese having a very high rate of deduplication ($\sim$75\%) and other such as Portuguese having a lower rate of deduplication ($\sim$40\%). This may be due to the biases of the embedding model or due to the kind of prompts submitted to the original dataset in different languages.

A table of the number of prompts filtered at each stage of our cleaning process can be found in Table~\ref{tab:cleaning_stages}.

\begin{table}[]
\begin{tabular}{|l|l|}
\hline
\textbf{Stage} & \textbf{Number of prompts} \\ \hline
Start & 1,000,000 \\ \hline
\begin{tabular}[c]{@{}l@{}}OpenAI Moderation \\ check\end{tabular} & 964,464 \\ \hline
\begin{tabular}[c]{@{}l@{}}Remove unknown \\ languages\end{tabular} & 936,468 \\ \hline
Remove anonymised data & 753,731 \\ \hline
\begin{tabular}[c]{@{}l@{}}Remove references \\ to models\end{tabular} & 735,390 \\ \hline
\begin{tabular}[c]{@{}l@{}}Language detection \\ confidence score \textgreater{}80\%\end{tabular} & 556,368 \\ \hline
\begin{tabular}[c]{@{}l@{}}Remove prompt plus \\ responses with more \\ than 512 tokens\end{tabular} & 513,011 \\ \hline
\begin{tabular}[c]{@{}l@{}}Random sampling \\ of 25,000 prompts\\ per language\end{tabular} & 157,873 \\ \hline
Fuzzy deduplication & 78,057 \\ \hline
\begin{tabular}[c]{@{}l@{}}Remove uncompleted/\\unanswered prompts\end{tabular} & 76,338 \\ \hline
\end{tabular}
\caption{Table describing the number of prompts after each cleaning stage.}
\label{tab:cleaning_stages}
\end{table}


We used these prompts to generate responses using an Azure OpenAI deployment of a state-of-the-art proprietary LLM, GPT4 (0125-Preview), with the generation temperature set to 0 and setting a maximum number of response tokens to be 2,048. 

Due to the fact that generating high quality responses for all of these prompts manually for each language would be prohibitively expensive, we decided to generate these responses using a state-of-the-art model. We hypothesize that using an LLM much larger - rumoured to be 1.8 trillion parameters~\cite{Schreiner_2023} - than nearly all other open source models to generate responses will lead to high quality responses that can then be used to improve existing open source models. When viewed in this way, this training can be viewed as a form of model distillation~\cite{bucilua2006model,hinton2015distilling}.

We finally removed any responses which GPT4 did not answer or was not able to complete within the 2,048 token limit. The number of prompts in our resultant Tagengo dataset can be found in Table~\ref{tab:cleaning_stages} and a breakdown of the prompts by language can be found in Fig~\ref{fig:language distribution}.

We share our dataset creation code and this training dataset on Huggingface at lightblue/tagengo-gpt4\footnote{\url{https://huggingface.co/datasets/lightblue/tagengo-gpt4}}.

\begin{table*}[]
\centering
\begin{tabular}{|l|l|l|l|l|l|}
\hline
 & \textbf{\begin{tabular}[c]{@{}l@{}}Llama 3 8B \\ Instruct\end{tabular}} & \textbf{\begin{tabular}[c]{@{}l@{}}Suzume 8B \\ multilingual\end{tabular}} & \textbf{\begin{tabular}[c]{@{}l@{}}Suzume 8B \\ Japanese\end{tabular}} & \textbf{\begin{tabular}[c]{@{}l@{}}Starling \\ 7B beta\end{tabular}} & \textbf{\begin{tabular}[c]{@{}l@{}}GPT3.5\\ Turbo\end{tabular}} \\ \hline
Chinese & - & 7.11 & - & 6.97 & 7.55 \\ \hline
French & - & 7.66 & - & 7.29 & 7.74 \\ \hline
German & - & 7.26 & - & 6.99 & 7.68 \\ \hline
Japanese & - & 6.56 & 6.24 & 6.22 & 7.84 \\ \hline
Russian & - & 8.19 & - & 8.28 & 7.94 \\ \hline
English & 7.98 & 7.73 & - & 7.92 & 8.26 \\ \hline
\end{tabular}
\caption{Average MT-Bench scores across 6 languages for each LLM evaluated.}
\label{tab:mtbench-results}
\end{table*}

\subsection{Training}

For training data, we add two more datasets to the Tagengo dataset which we regard as high quality chat datasets. The first is the Megagon Instruction dataset~\cite{megagonlabs_instruction_ja}, a manually annotated dataset of 669 Japanese prompt-response pairs. The second is the 6k GPT4 subset of the ShareGPT dataset\footnote{\url{https://huggingface.co/datasets/openchat/openchat_sharegpt4_dataset/blob/main/sharegpt_gpt4.json}}, which has a majority of prompts in English but also includes responses in other languages. We combined and randomly shuffled these three datasets to use as a 83,213 prompt-response pair training dataset for the multilingual model.

We used our training data to train a Llama 3 8B Instruct model\footnote{\url{https://huggingface.co/meta-llama/Meta-Llama-3-8B-Instruct}} with the Axolotl LLM training package\footnote{\url{https://github.com/OpenAccess-AI-Collective/axolotl}}. We trained for one epoch using full fine tuning, using sample packing~\cite{brown2020language} and a context length of 8,096. We name this model Suzume 8B multilingual and the full training configuration for this model can be found on our model card\footnote{\url{https://huggingface.co/lightblue/suzume-llama-3-8B-multilingual}}.

We also prepared a subset of the above three datasets that only included Japanese data from each dataset, which amounted to 3,318 prompt-response pairs. This was prepared to isolate the effect of monolingual training compared to multilingual training on our data. We trained our model in the same manner as the multilingual model with the name Suzume 8B Japanese. Full details for how the training was conducted can be found on our model card\footnote{\url{https://huggingface.co/lightblue/suzume-llama-3-8B-japanese}}.

\subsection{Evaluation}

We tested our models by using a forked version of the original MT-Bench evaluation suite~\cite{zheng2024judging}. The MT-Bench evaluation benchmark is a set of prompts and responses in English that cover 8 broad categories of prompts: writing, roleplay, extraction, reasoning, math, coding, STEM knowledge, and humanities knowledge. Responses to these prompts are generated using an LLM, and those responses are then evaluated using an evaluation model such as GPT4.

We added publicly available translated versions of the original MT-Bench dataset in Chinese, French, German, Japanese, and Russian that had been human-verified by a native speaker of that language. 

Note that the Russian translation did not contain reference answers for the math, coding, and reasoning questions, so our evaluation does not include math, coding, and reasoning problems in Russian.

Finally, we added the phrase ``Your evaluation should also consider whether the prompt responded in the correct language and the fluency and naturalness of this response.'' to the original MT-Bench evaluation criteria to ensure that the LLM judge would not simply evaluate factually correct responses in English to non-English prompts as correct. We conducted these evaluations using the ``gpt-4-turbo model'' from OpenAI as the judge LLM.

We make our evaluation code freely available online\footnote{\url{https://github.com/Peter-Devine/multilingual_mt_bench/tree/main/fastchat/llm_judge}}.

As baselines, we also evaluate the original Llama 3 8B Instruct model~\cite{llama3modelcard}, GPT3.5-Turbo~\cite{ouyang2022training}, and the Starling 7B Beta~\cite{starling2023} which is the highest rated similarly sized multilingual model on the Chatbot Arena leaderboard~\cite{chiang2024chatbot} and has been trained on the ShareGPT dataset amongst other data.

\section{Results}

The MT-Bench scores for each model evaluated can be found in Table~\ref{tab:mtbench-results}.

We first found that we were able to train Llama 3 8B Instruct to output responses in the same language as the prompt. This means that we achieved our base objective of enabling a monolingual model (Llama 3) to be able to output multilingual chat.

Secondly, English performance of the multilingual trained model only dropped slightly compared to the base Llama 3 8B Instruct model. This indicates that English chat performance does not considerably drop even when training on a majority of non-English data.

Thirdly, we found that the multilingual trained model performs better compared to the Starling 7B Beta across 5 out of 6 non-English languages tested. However, also we found that our multilingual model achieved lower evaluation scores compared to the proprietary GPT3.5 on 5 of 6 non-English languages. This indicates that our model has achieved state-of-the-art performance in multilingual chat for open-source models of its size, but has not achieved state-of-the-art performance more generally.

Finally, the Suzume 8B multilingual model achieves higher MT-Bench scores on the Japanese benchmark compared to the Suzume 8B Japanese model, indicating that transfer learning from training on other languages is beneficial for training even monolingual models outside of English.

\section{Discussion}

Our results indicate the need for large, high quality, multilingual datasets when training multilingual models. We find that with such a dataset, we can train a state-of-the-art monolingual model such as Llama 3 to achieve state-of-the-art multilingual performance.

We also found that training on additional non-Japanese data improves the performance of our LLM on Japanese benchmarks when compared to training solely on Japanese data, indicating that there is a collective improvement effect between languages when training using multilingual data. This adds to the body of work that indicates that training on multiple languages enables the LLM to better generalise to other languages~\cite{nguyen2017transfer,schuster2018cross}. This suggests that generating an even larger, more diverse dataset in the future could further aid the performance of LLMs on low-resource languages.

\section{Future Work}
\label{sec:future}

Our work could be built upon and improved in the following ways.

Our training dataset mainly consisted of single prompt-response pairs, but many chats between users and LLMs extend beyond a single conversation turn. Therefore, future work could include creating a dataset that contains multiple turns of conversation, with the prompts either generated by humans or by high quality LLMs.

Future work could also include adding more languages to our dataset. Our dataset only included 74 languages, and crucially omits any languages in the Niger–Congo language family, one of the most diverse language families in the world~\cite{good2017niger}. Therefore, future work could involve sampling initial prompts from a wider range of sources (possibly by advertising free chatbot access to people in areas with many speakers of underrepresented languages) and generating responses based on these prompts. This would help to both improve an LLMs linguistic understanding of these low-resource languages as well as improve their understanding of the topics and questions that people from that language and culture may ask.

Finally, future work could include generating preference data, such as was done in English in the Nectar dataset~\cite{starling2023}, for use with contrastive learning techniques such as Direct Preference Optimisation~\cite{rafailov2024direct} and Odds Ratio Preference Optimisation~\cite{hong2024reference}. These techniques have been shown to further improve the accuracy of LLMs, suggesting that training using these techniques may also improve the performance of LLMs in multilingual chat.

\section{Conclusion}

In this study, we successfully trained a state-of-the-art monolingual Llama 3 LLM to chat multilingually using a new, diverse dataset comprising over 70k human-generated prompts in 74 languages paired with high-quality synthetic responses. Our multilingual model showcased superior performance across multiple languages compared to similar-sized open-source models on various chat benchmarks. Interestingly, training using a multilingual dataset also enhanced the performance on specific monolingual tasks, implying beneficial cross-linguistic transfer effects. The outcomes underlines the importance of using rich, diverse multilingual data for improving the capabilities of LLMs in global, multilingual applications.

\section*{Limitations}

The three main limitations of this paper concern our prompt diversity, our data generation methodology, and our model evaluation methodology. 

Firstly, as stated in Section~\ref{sec:future}, our training data has a paucity of low-resource languages represented within it. While we try to focus on non-English data in our work by sampling a maximum of 25,000 prompts per language, this still does not counteract the fact that the prompts in the LMSYS-Chat-1M dataset~\cite{zheng2023lmsyschat1m} are disproportionately from a small set of languages. These prompts are collected from users on the Chatbot Arena LLM demo site, meaning that the speakers of low-resource languages may be too few, unable, unaware, or unwilling to talk to an LLM chatbot in their native language. This means that current open source LLMs will continue to have lower performance on low-resource languages if this problem is not resolved.

Secondly, we generate our responses to prompts using GPT4, which means that all training data will be in the worldview and within the domain of knowledge that GPT4 exhibits. This biases the model as many LLMs have been shown to have both political~\cite{feng2023pretraining} and cultural biases~\cite{cao2023assessing} in the text they generate, meaning that what may be deemed acceptable by one user may not be deemed acceptable by another. Moreover, while GPT4 is state-of-the-art and has been shown to generate more accurate information compared to previous models~\cite{achiam2023gpt}, it is still capable of generating incorrect data in response to a prompt, meaning that our training data may contain incorrect statements or otherwise inaccurate data.

Finally, our evaluation methodology is biased by the fact that our 6 evaluation languages are all languages that are within the top 10 most popular languages in our training data. This means that our evaluation does not consider the performance of our models on low resource languages, limiting the usefulness of our results to speakers of low resource languages.



\bibliography{custom}
\bibliographystyle{acl_natbib}

\appendix

\section{Appendix: Full Language Counts}
\label{sec:appendix}

\begin{table*}[]
\centering
\begin{tabular}{|l|r|l|l|r|}
\hline
\textbf{Language} & \textbf{Number of examples} &  & \textbf{Language} & \multicolumn{1}{l|}{\textbf{Number of examples}} \\ \hline
English & 15,404 &  & Croatian & 11 \\ \hline
Portuguese & 12,412 &  & Lithuanian & 11 \\ \hline
Spanish & 8,098 &  & Slovenian & 10 \\ \hline
Russian & 8,005 &  & Mongolian & 6 \\ \hline
Italian & 6,874 &  & Basque & 6 \\ \hline
German & 5,430 &  & Serbian & 6 \\ \hline
Chinese & 5,309 &  & Icelandic & 5 \\ \hline
French & 5,163 &  & Macedonian & 5 \\ \hline
Japanese & 2,482 &  & Albanian & 5 \\ \hline
Korean & 1,604 &  & Malay & 5 \\ \hline
Polish & 1,029 &  & Sinhala & 4 \\ \hline
Arabic & 783 &  & Latin & 4 \\ \hline
Vietnamese & 428 &  & Tamil & 4 \\ \hline
Turkish & 401 &  & Azerbaijani & 4 \\ \hline
Dutch & 375 &  & Amharic & 3 \\ \hline
Ukrainian & 322 &  & Armenian & 3 \\ \hline
Greek & 293 &  & Urdu & 3 \\ \hline
Swedish & 241 &  & Afrikaans & 2 \\ \hline
Indonesian & 234 &  & Kazakh & 2 \\ \hline
Hungarian & 210 &  & Belarusian & 2 \\ \hline
Persian & 183 &  & Waray & 2 \\ \hline
Czech & 176 &  & Yiddish & 2 \\ \hline
Thai & 131 &  & Uyghur & 2 \\ \hline
Hebrew & 117 &  & Burmese & 2 \\ \hline
Finnish & 88 &  & Tibetan & 1 \\ \hline
Catalan & 73 &  & Turkmen & 1 \\ \hline
Romanian & 71 &  & Breton & 1 \\ \hline
Danish & 64 &  & Lao & 1 \\ \hline
Bulgarian & 56 &  & Georgian & 1 \\ \hline
Bangla & 28 &  & Sanskrit & 1 \\ \hline
Norwegian & 25 &  & Khmer & 1 \\ \hline
Hindi & 20 &  & Kannada & 1 \\ \hline
Latvian & 20 &  & Luxembourgish & 1 \\ \hline
Estonian & 18 &  & Odia & 1 \\ \hline
Esperanto & 17 &  & Marathi & 1 \\ \hline
Slovak & 17 &  & Malayalam & 1 \\ \hline
Tagalog & 15 &  & Uzbek & 1 \\ \hline
\end{tabular}
\caption{The number of prompt-response examples in our Tagengo dataset for each language}
\label{tab:dataset-languages}
\end{table*}
\end{document}